\documentclass[letterpaper, 10 pt, conference]{ieeeconf}
\IEEEoverridecommandlockouts                           
\usepackage{amsmath,amssymb,amsfonts}
\usepackage{graphicx}
\usepackage{xcolor}
\usepackage{array}
\usepackage{tabularx}
\usepackage{booktabs}
\usepackage{csquotes}
\usepackage{textcomp}
\usepackage{url}
\usepackage{verbatim}
\usepackage{algorithm}
\usepackage{algpseudocode}
\usepackage{listings}
\usepackage{wrapfig}
\usepackage{adjustbox}
\usepackage{subfigure}  
\usepackage{comment}
\usepackage{afterpage}
\usepackage{lipsum}

\begin{document}

\title{\LARGE \bf DiffusionRL: Efficient Training of Diffusion Policies for Robotic Grasping Using RL-Adapted Large-Scale Datasets}

\author{Maria Makarova$^{1}$, Qian Liu$^{2}$ and Dzmitry Tsetserukou$^{1}$
\thanks{$^{1}$The authors are with the Intelligent Space Robotics Laboratory, Center for Digital Engineering, Skolkovo Institute of Science and Technology (Skoltech), 121205 Moscow, Russia. 
{\tt \small $\{$maria.makarova2, d.tsetserukou$\}$@skoltech.ru}}
\thanks{$^{2}$ Qian Liu is with the Department of Computer
Science and Technology, Dalian University of Technology, China. {\tt\small qianliu@dlut.edu.cn}}
}

\maketitle

%===============================================================================
\begin{abstract}
    %Diffusion models have been successfully applied in areas such as image, video, and audio generation. Recent works show their promise for sequential decision-making and dexterous manipulation, leveraging their ability to model complex action distributions. However, challenges persist due to the data limitations and scenario-specific adaptation needs. In this paper, we address these challenges by proposing an optimized approach to training diffusion policies using large, pre-built datasets that are enhanced using Reinforcement Learning (RL). Our end-to-end pipeline leverages RL-based enhancement of the DexGraspNet dataset, lightweight diffusion policy training on a dexterous manipulation task for a five-fingered robotic hand, and a pose sampling algorithm for validation. The pipeline provided a high success rate of 80\% for three DexGraspNet objects in random poses. By eliminating manual data collection, our approach lowers barriers to adopting diffusion models in robotics, enhancing generalization and robustness for real-world applications. 
    Diffusion models have been successfully applied in areas such as image, video, and audio generation. Recent works show their promise for sequential decision-making and dexterous manipulation, leveraging their ability to model complex action distributions. However, challenges remain due to data limitations and the need for adaptation to specific scenarios. In this paper, we address these challenges by training diffusion policies using large, pre-built datasets enhanced and adapted to the desired scenarios using reinforcement learning (RL). Our pipeline leverages RL-based adaptation of the DexGraspNet dataset, diffusion policy training on a dexterous manipulation task for a five-fingered robotic hand, and an algorithm for sampling realistic object poses for validation. The pipeline achieved a high success rate (80\%) for three DexGraspNet objects in random poses of the robotic hand and object. By eliminating manual data collection, our approach lowers barriers to adopting diffusion models in robotics, enhancing generalization and robustness for real-world applications. 
    \end{abstract}
    %However, challenges remain due to data limitations and the need for adaptation to specific scenarios. In this paper, we address these challenges by proposing an optimized approach to training diffusion policies using large, pre-built datasets that are enhanced and adapted to the desired scenarios using reinforcement learning (RL). Our end-to-end pipeline leverages RL-based adaptation of the DexGraspNet dataset, lightweight diffusion policy training on a dexterous manipulation task for a five-fingered robotic hand, and an algorithm for sampling realistic object poses for validation. The pipeline achieved a high success rate (80\%) for three DexGraspNet objects in random poses with gravity taken into account. By eliminating manual data collection, our approach lowers barriers to adopting diffusion models in robotics, enhancing generalization and robustness for real-world applications. 

%\keywords{Diffusion Policy, Reinforcement Learning, Dexterous Manipulation} 

\section{Introduction}

Diffusion models have proven to be a powerful tool in the field of generative artificial intelligence successfully applied in image synthesis, video and audio generation \cite{grasp_1,grasp_2, grasp_3,grasp_4,grasp_survey}. Using an iterative denoising approach, these models learn to invert a diffusion process, transforming random noise into sophisticated, high-quality samples.
 
Reinforcement Learning (RL) and Imitation Learning (IL) have become particularly popular in the field of robot learning for the tasks of perceiving the environment and making decisions to perform actions in recent years \cite{grasp_7}. But RL approach is highly dependent on the correct tuning of hyperparameters \cite{eimer2023hyperparametersreinforcementlearningtune}, and effective IL training requires a large amount of diverse high-quality data \cite{belkhale2023dataqualityimitationlearning}. Also, the multimodal nature of complex robot tasks hinders the construction of stable control. More recently, researchers have begun to integrate an approach in the form of diffusion policy learning into the field of robotics as well. 

The concept of Diffusion policy was first introduced by Chi et al. \cite{grasp_8}. The diffusion process has been applied to robot action sequence generation since such models are able to capture the complex multimodal distributions that are characteristic of many robotics tasks, as mentioned above. 
Diffusion models have found applications in constrained motion planning \cite{grasp152, grasp153}, human-robot interaction \cite{grasp161}, mobile manipulation \cite{grasp61, grasp62} and navigation \cite{grasp64, grasp65}. 

\begin{figure}[h]
    \centering
      \includegraphics[width=1.0\linewidth]{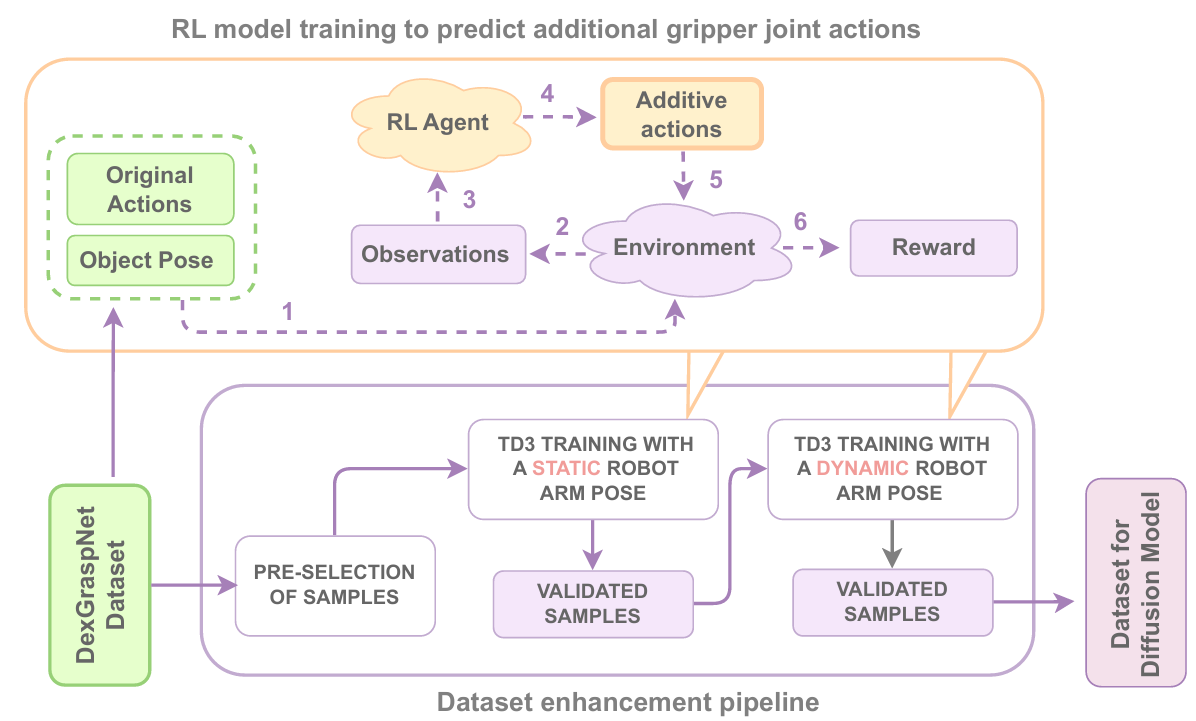}
\caption{RL-based Dataset Enhancement Pipeline. The numbers indicate the data flow sequence during the training of the RL agent: from DexGraspNet, the data on the object pose and actions for the gripper are sent to the Environment (1), from where the Observations vector is received by the RL agent (2, 3), after which it predicts additional gripper actions for the Environment (4, 5), where the final Reward is calculated to validate the success of the object grasping process (6).}{\label{fig:scheme}}
\vspace{-0.4cm}
\end{figure}

They have also begun to be applied in dexterous robotic manipulation, which is one of the most demanded tasks. DexDiffuser \cite{grasp_158} introduces DexSampler, which generates grasping poses based on the point cloud of a 3D object using the diffusion process. GraspGF \cite{grasp_158} is an algorithm that uses a generative model as a primitive hand-object-conditional policy. DexGrasp-Diffusion \cite{grasp_159} combines multi-dexterous hand grasp estimation with open-vocabulary setting to filter physically plausible functional grasps based on object affordances.

In other works \cite{grasp_59,grasp_118_main} authors utilize video demonstrations and Vision-Language-Action (VLA) models. In the latter work, the VLA provides high-level planning through language commands, while the diffusion model handles low-level interactions that ensure accuracy and robustness. The VLA guides the robot to the target object, after which the diffusion model implements ADAPT Hand 2 \cite{adapthand} grasping actions. The authors demonstrated that the model-switching approach yields more than 80\% success rate compared to less than 40\% using the VLA model alone. However, the diffusion model training dataset consisted of 30-40 samples for each scenario, which does not allow efficient generalization to other objects.

Therefore, we have decided to simplify the integration of this approach and provide the ability to train diffusion models based on large pre-built datasets containing various examples of object poses and actions for grasping. As such a dataset, we adopted the DexGraspNet \cite{wang2023dexgraspnetlargescaleroboticdexterous} dataset for ShadowHand \cite{shadowrobot2024} with 24 degrees of freedom, containing 1.32 million capture examples for 5355 objects, covering more than 133 object categories and containing more than 200 different captures for each object instance. But this dataset cannot always be successfully applied to training on individual tasks. For example, the authors of DexDiffuser \cite{grasp_158} also described a large percentage of inaccurate samples, reaching more than 50\%.
%But this dataset is not completely clean, for example, the authors of DexDiffuser \citep{weng2024dexdiffusergeneratingdexterousgrasps} also described a large percentage of inaccurate samples from it, reaching more than 50\%. 
%We developed an end-to-end pipeline to collect a customized dataset in a ManiSkill \citep{maniskill3gpuparallelizedrobotics} environment based on DexGraspNet using an RL agent that improves the accuracy and stability of object grasping, and a pipeline to train a diffusion model based on the collected data and validate it on randomly generated object poses computed from statistics from the original dataset.

We developed an end-to-end pipeline that allows to efficiently train lightweight diffusion policies on existing datasets customized for specific tasks and environmental conditions using an RL agent, significantly reducing the need for manual data collection in complex manipulation tasks and increasing scalability.
The main contributions of the presented work
can be summarized as follows:
\begin{itemize}
\item A pipeline was developed to enhance the DexGraspNet samples using a lightweight RL model in a changing robot pose scenario. The resulting high quality dataset was used to train a diffusion model in the dexterous grasping task with the ShadowHand.
\item To validate the success of the training, an algorithm was developed to generate random object poses based on the statistics of the original DexGraspNet dataset.
\item A compact diffusion model was trained on three different objects, and its performance in grasping objects in random scenarios reached a high value of about 80\%.
\end{itemize}

Our method can be used in a large set of real-world complex robotic manipulation tasks, as it allows to train lightweight diffusion models on large-scale pre-built datasets, adapting them to specific conditions without the need to collect extensive data manually.

%=======================================================================

\section{Methods}\label{chapter_grasp}

\subsection{Dataset Enhancement with Reinforcement Learning.}\label{dataset_chapter}{
%The training of a diffusion model for grasping complex objects located in the workspace of the UR10e robot is based on the DexGraspNet dataset. Simply transferring this dataset to the simulation environment to train the diffusion model is a problematic. The application of some object and hand poses depends on a specific hand position, as shown in Fig. \ref{fig:pipek}(c). If the hand pose is changed, the stability of the object during grasping will be compromised. Also the robotic hand models in the dataset and the simulated environment have some discrepancies, resulting in instability of successful grasping.
%Therefore, to effectively train the diffusion model, an RL agent was built to \textbf{predict additional grasping actions} in addition to those taken from the original dataset. These  actions are designed to provide reliable and stable object grasping in different poses of the UR10e robotic arm. 
The training of a diffusion model for grasping complex objects located in the workspace of the UR10e robot is based on the DexGraspNet dataset. The simple transfer of this dataset to the simulation environment for the training of the diffusion model is problematic. The implementation of some object and hand poses is contingent on a particular hand position, as illustrated in Fig. \ref{fig:pipek}(a). A variation in the hand pose will result in a compromised object's stability during grasping. Moreover, the ShadowHand models in the dataset and the simulated environment exhibit discrepancies, leading to instability in successful grasping.

Therefore, in order to effectively train the diffusion model, an RL agent was built to \textbf{predict additional grasping actions} in addition to those taken from the original dataset. The purpose of these actions is to ensure the reliability and stability of object grasping in different poses of the robotic hand. 

The process of enhancing the dataset is shown in Fig. \ref{fig:scheme}. At first, in the Isaac Sim environment \cite{makoviychuk2021isaac} used by the authors of the original DexGraspNet, the direction of gravity varied from -25$^{\circ}$ to 25$^{\circ}$ relative to the vertical. This was done to exclude unreliable examples of grips in which a small change in the direction of gravity caused the object to fall out.
Then, in the ManiSkill environment \cite{maniskill3gpuparallelizedrobotics}, the RL agent was trained to predict additional actions in a stationary pose of the UR10e robot with the ShadowHand positioned vertically. This approach was used to determine which samples from the dataset could not be improved by the RL agent, even in the simplest cases. Finally, samples from the resulted subset were used to train the RL agent to predict additional actions when the position of the robot was changed randomly at the beginning of each episode. 

As a result, a new dataset was obtained for training the diffusion model.

\vspace{0.2cm}
\textbf{Reinforcement Learning Agent.}\label{rl_grasp_theory}

After positioning the robot at the beginning of each episode in the ManiSkill environment, the object's position is recalculated based on the original DexGraspNet dataset by calculating the inverse positional and rotational transformations from the robotic hand. The object is placed in the calculated area under zero gravity conditions. Then, the movement of the ShadowHand joints from the DexGraspNet dataset is implemented, followed by additional movement of the robotic hand joints predicted by the RL agent. Then gravity is turned on and a gripping check is performed: if the object has not fallen out of the robotic hand, the check is successful, and vice versa. A scheme of the described process is shown in Fig. \ref{fig:pipek}(b). Training was based on the Twin Delayed DDPG (TD3) algorithm \cite{lillicrap2019continuouscontroldeepreinforcement}. The Critic and Actor models were built as lightweight architectures (Fig. \ref{fig:unet}(a)), each consisting of four linear layers, since the speed of trajectory calculation is one of the key characteristics in real-world scenarios. 

\begin{figure}[h]
  \centering
  %\subfigure[]{
%\includegraphics[width=0.9\linewidth]{pictures_grasp/grasp-pipeline.png}}
  \subfigure[]{
\includegraphics[width=0.3\linewidth]{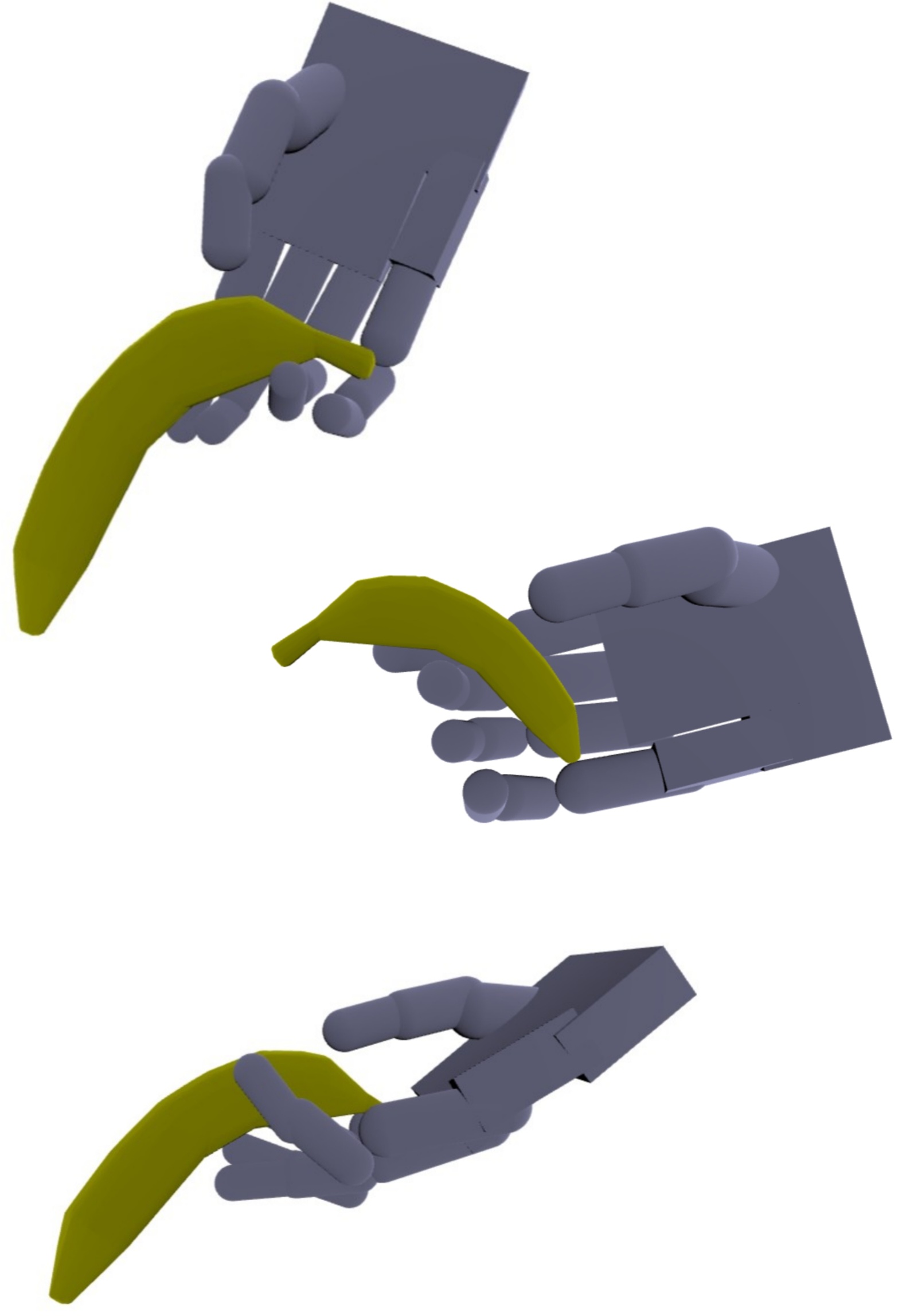}
  }
    \subfigure[]{
\includegraphics[width=0.35\linewidth]{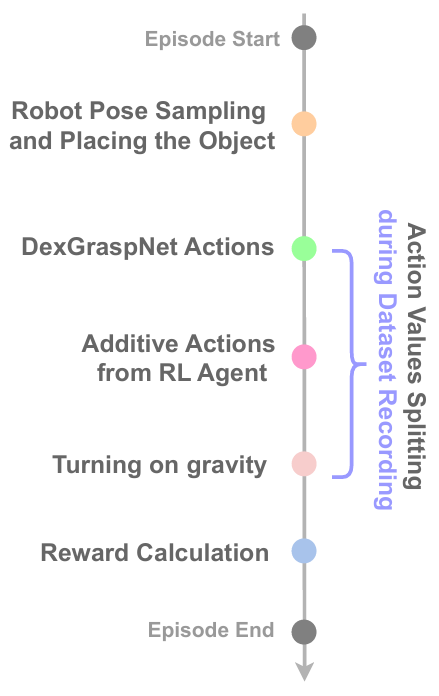}} 
\caption{(a) Examples of inappropriate DexGraspNet samples that were excluded during the Pre-selection stage; (b) Environmental Timestamps during RL-agent Training and Dataset Recording.}\label{fig:pipek}
\end{figure}
\vspace{-0.35cm}

\subsection{Diffusion Model.}\label{diffusion_theory} 

The environment differs from the one previously described for the RL agent in that information about the original robotic hand joint actions from DexGraspNet is no longer used. The diffusion model is trained using enhanced trajectories obtained earlier using an RL agent and directly predicts the sequence of ShadowHand joint actions. In validation mode, the object is placed in a random pose using an algorithm that will be described below.

\begin{figure*}[ht]
  \centering
    \subfigure[]{
  \includegraphics[width=0.25\linewidth]{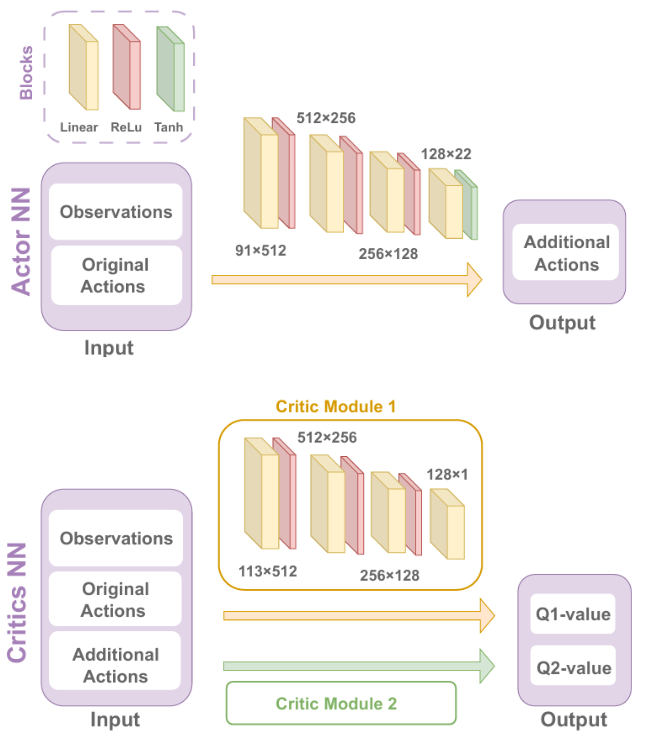}} 
  \hfill
  \subfigure[]{
  \includegraphics[width=0.7\linewidth]{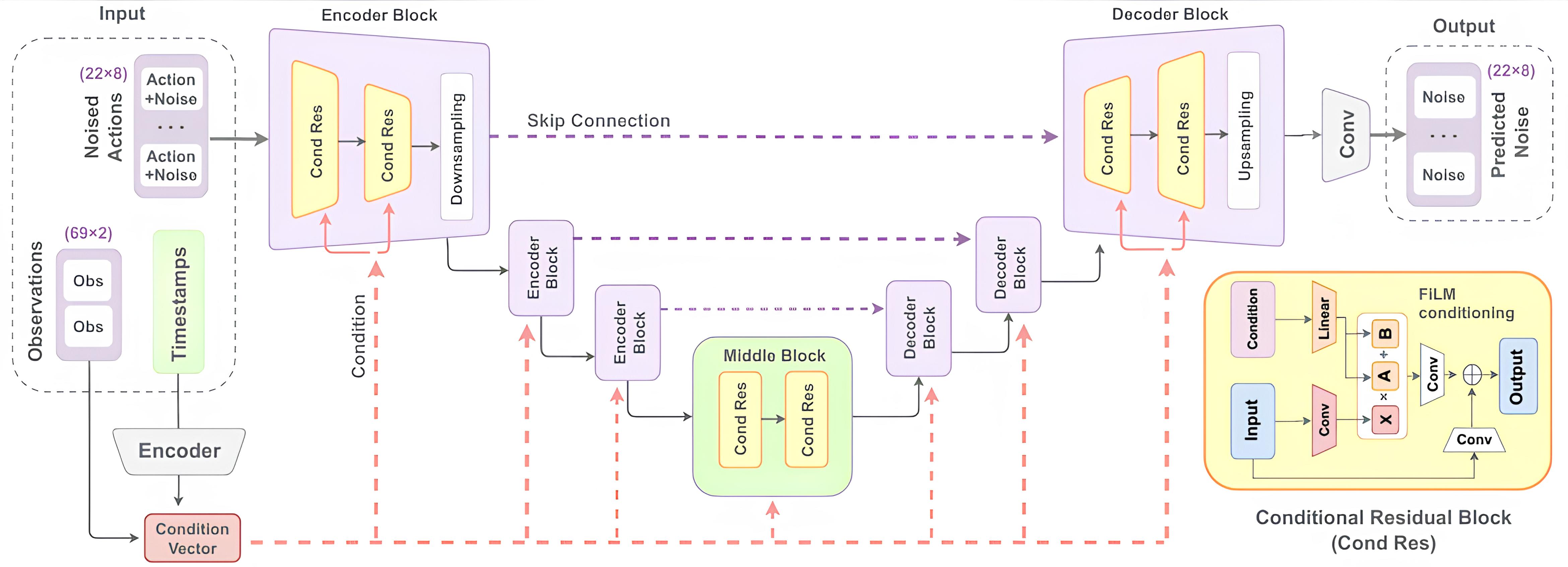}}
 
\caption{(a) Lightweight Architectures for Actor and Critic; (b) Diffusion Model Conditional U-net Architecture. Conditional Residual Blocks are highlighted in yellow.}\label{fig:unet}
\vspace{-0.5cm}
\end{figure*}

As in the original work on Diffusion policy \cite{grasp_8}, the samples of the dataset have been transformed. The model architecture was constructed as a Conditional U-net (Fig. \ref{fig:unet}(b)), as proposed in the original paper \cite{ronneberger2015unetconvolutionalnetworksbiomedical,grasp_8}. The DDPM model \cite{ho2020denoisingdiffusionprobabilisticmodels} uses the observation history of the 2 previous steps, predicts 8 action vectors, and implements half of them in the environment. For vectors that precede the initial vectors in the episode, the context is duplicated by the initial vectors. Similarly, for the final vectors, the context is duplicated by them. An illustration for a recalculated sample is shown in Fig. \ref{fig:recalculated}.

\begin{figure}[h]
    \centering
      \includegraphics[width=0.6\linewidth]{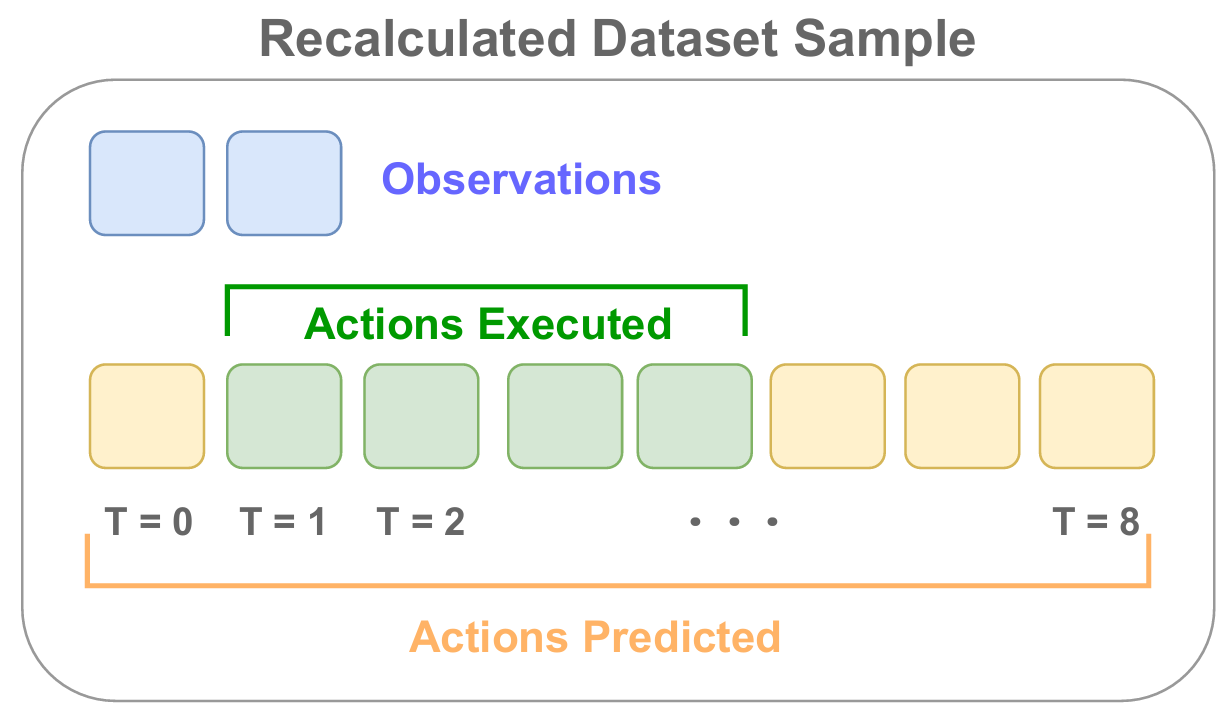}
\caption{Recalculated dataset sample for Diffusion Model training.}{\label{fig:recalculated}}
\vspace{-0.4cm}
\end{figure}

\section{Experiments}

\textbf{Environment description.}

The environment built in ManiSkill contains a UR10e manipulator with Shadow Dexterous Hand. The UR10e robot has 6 joints, the ShadowHand has 24 joints. At the beginning of each episode, the robot's position is set for all 6 joints of the UR10e ($i = {1,...,6}$) and for the 2 joints of the ShadowHand connecting the palm to the wrist and the wrist to the forearm ($i = 7,8$). The values of the joint angles vary by $A_i$ from the initial value of joint $S_i$, characterizing its values for the static position of the robot (the ShadowHand palm is in a vertical position). These values for each of the 8 joints are: 
$[A_1, \ldots, A_8] = [\pi/2; \pi/12; \pi/12; \pi/12; \pi/8; \pi/8; \pi/8; \pi/8]$; 
$[S_1, \ldots, S_8] = [0; -\pi/2; 3\pi/4; -3\pi/4; \pi/2; -\pi; 0; 0].$
The formula for calculating the resulting values of joint angles is $R_i = S_i + w_i *A_i$, where $w_i$ is a random number from $[-1;1]$.

The \textit{Action} vector for the RL agent is of size 22 and is used to control the robotic hand joints except for the two previously described. The \textit{Observation} vector for the agent consists of all 30 values of the angles of the robotic arm and hand joints, the object position (translation vector and rotation quaternion) relative to the ShadowHand base, the initial global position of the object at the beginning of the episode, the current global position of the object, the global position of the robotic hand base, and the global translation vectors of the five joints of the upper phalanges of the ShadowHand. In addition to these values, the RL agent receives 22 values characterizing the values of the ShadowHand joint angles from DexGraspNet. Thus, the final \textit{Observation} vector has a size of 91. The \textit{Reward} was chosen to be sparse and calculated as follows: \textit{Reward} $ = -1$ if the vertical coordinate of the object deviated from the coordinate set at the beginning of the episode by an amount greater than $d = 0.025$ after gravity was turned on, and \textit{Reward} $ =0$ otherwise. The choice of the value of $d$ was based on the high stability with varied initial conditions. All objects of the same type (Banana, Bottle and Camera) are considered at a single scale, which was calculated as the average value for that object in the original dataset.

\vspace{0.2cm}

\textbf{Selection of a representative subset of DexGraspNet.}
We validated our approach by selecting a limited set of objects from DexGraspNet: a Banana, a Bottle, and a Camera (Fig. \ref{ruki}). This selection was based on the need to cover key categories of shapes that pose challenges for planning grips. The Banana represents a class of elongated curved shapes that lack axial symmetry and require precise grips. The Bottle, as an axially symmetric object with complex topology, tests the ability to generate stable grips for shapes with variable cross-sections. The Camera, being a complex composite mesh without dominant symmetry, serves as a  test for handling irregular shapes with protruding elements. This set allows to focus on the depth of analysis and verify the method's ability to generalize solutions for diverse geometric challenges at the proof-of-concept stage.

\subsection{RL-based Dataset Enhancement.}

As described in Section \ref{dataset_chapter}, after pre-selection of samples from DexGraspNet, the RL agent was trained to predict additional actions for the ShadowHand joints when the UR10e is static (with the ShadowHand palm upright). The training results are summarized in Fig. \ref{fig:rl_diffusion_plots}(a, b). As can be seen, the agent was not completely successful because not all samples from the dataset could be effectively improved with RL. Some samples were too imprecise and additional actions did not help in object grasping. Therefore, they were also excluded from the dataset. The remaining samples were used to train the RL agent with a random pose of the robot at the beginning of each episode. The training results are shown in Fig. \ref{fig:rl_diffusion_plots}(c, d).
\begin{figure}[h!]
  \centering
    \subfigure[Mean Reward. Static robot pose.]{\includegraphics[width=0.485\linewidth]{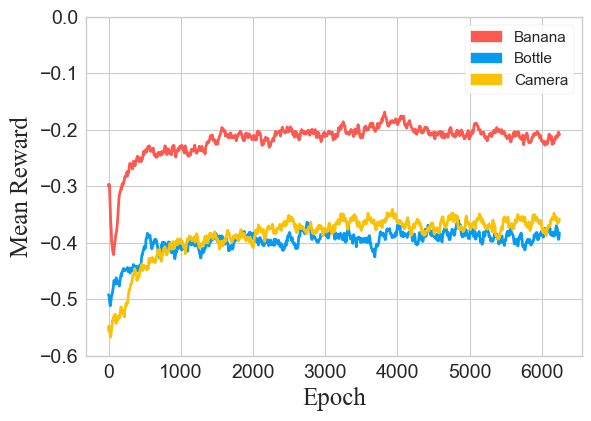}}
  \subfigure[Success Rate. Static robot pose.]{ \includegraphics[width=0.485\linewidth]{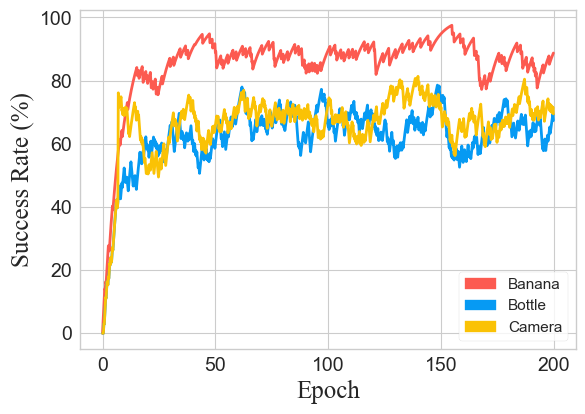}}
    \subfigure[Mean Reward. Random robot pose.]{\includegraphics[width=0.485\linewidth]{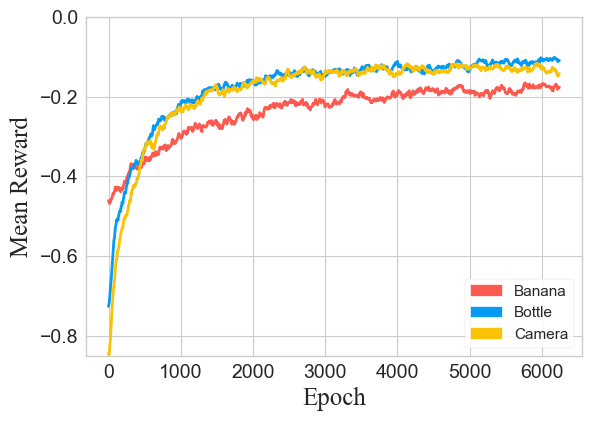}}
   \subfigure[Success Rate. Random robot pose.]{\includegraphics[width=0.485\linewidth]{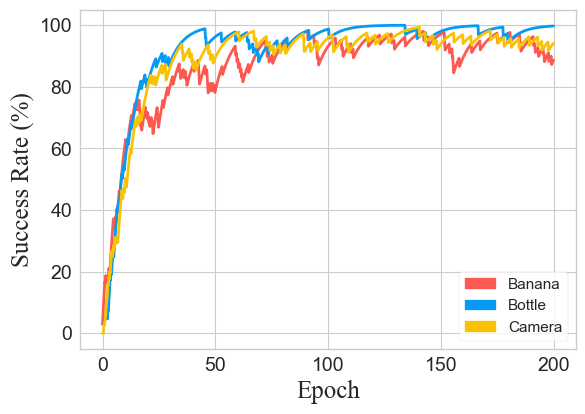}}

   \caption{Mean Reward and Mean Success Rate values during training an RL agent with static and random robot pose to predict additional ShadowHand joint actions. Success Rate means mean percentage of transitions that have a success reward during validation. Training with random robot pose was performed on a subset of the data selected after training with static robot pose.}\label{fig:rl_diffusion_plots}
   
   \vspace{-0.25cm}
\end{figure}

\begin{figure}[h!]
  \centering
  \subfigure[]
  {\includegraphics[width=0.75\linewidth]{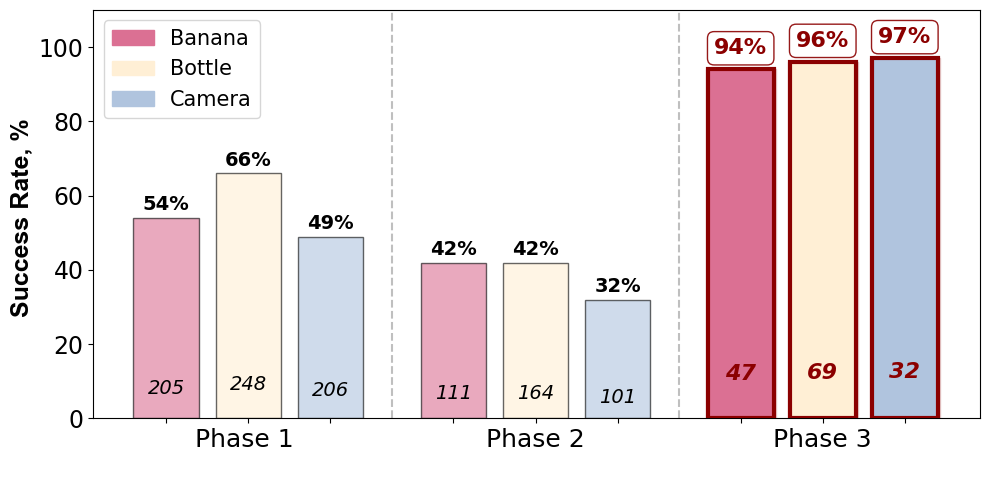}}
   \subfigure[] 
    {\includegraphics[width=0.75\linewidth]{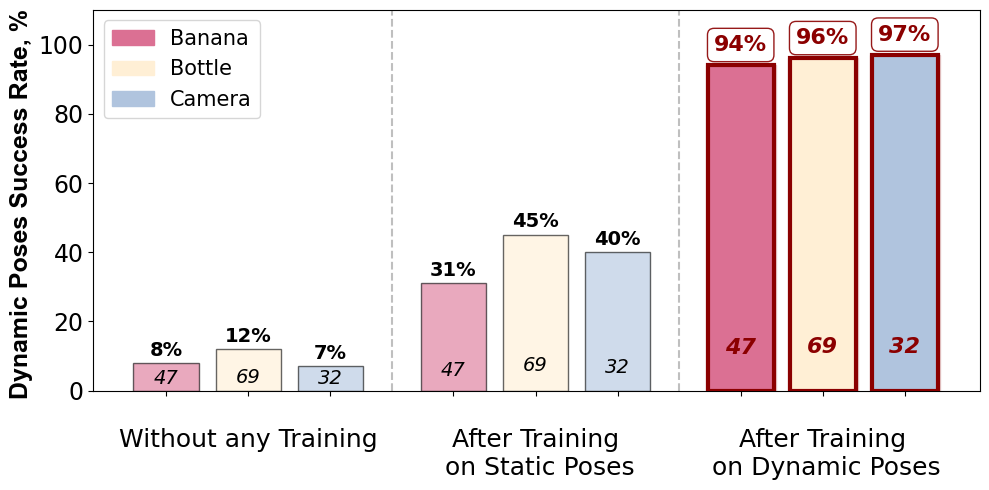}}
     \caption{(a) The ratio of successful samples in datasets after Phase 1 (Pre-selection), Phase 2 (training the RL agent with a static robot pose), and Phase 3 (training the RL agent with a random robot pose). The numbers inside the graphs indicate the number of unique samples from DexGraspNet taken for each of the phases; 
    (b) The ratio of successful samples in a scenario with a random robot pose in the absence of any training, after RL-agent training in a static pose, and after training in a random pose. Samples were taken from those used during Phase 3.
}\label{fig:ablation}
 \vspace{-0.5cm}
\end{figure}

The results demonstrate that the RL agent was successfully trained and achieved convergence close to 90-100\% success rate. Training of the RL-agent in each of the experiments occurred over 200 epochs of 75 episodes each.

A dataset of 2000 episodes was then recorded for each object. The actions for the ShadowHand joints were limited to $25\times10^{-4}$ for each time step for smoother motion, which is important for the robustness of the diffusion model training (Fig. \ref{fig:pipek}(b)). Information about the proportions of selected samples after each stage of the pipeline is shown in Fig. \ref{fig:ablation}(a).
The dataset now contains high-quality samples and is suitable for training the diffusion model.

To show that the described pipeline effectively enhances the dataset samples through RL agent training rather than simply selecting a subset step by step, we conducted a reverse experiment. Using the DexGraspNet samples used in the final enhanced dataset, we conducted experiments on object grasping with a random initial pose of the robot, first without any training in the ManiSkill environment and then after training the RL agent only with a static robot pose. As shown in Fig. \ref{fig:ablation}(b), the RL agent enabled the robot to reliably grasp objects of various shapes, regardless of their position or the direction of gravity, only at the final stage of the pipeline. 

\subsection{Algorithm for Sampling Object Poses.}\label{sampling_grasp}

\begin{figure}[h]
  \centering
  \subfigure[]{\includegraphics[width=0.24\linewidth]{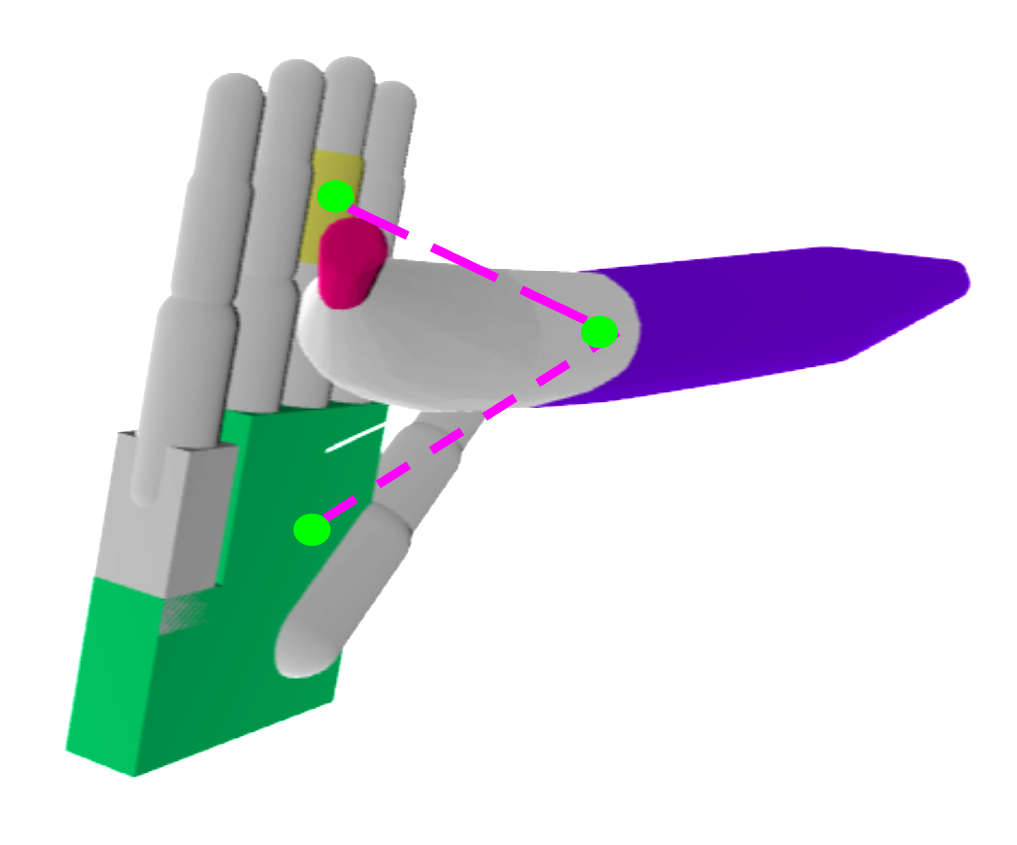}}
  \subfigure[]{\includegraphics[width=0.25\linewidth]{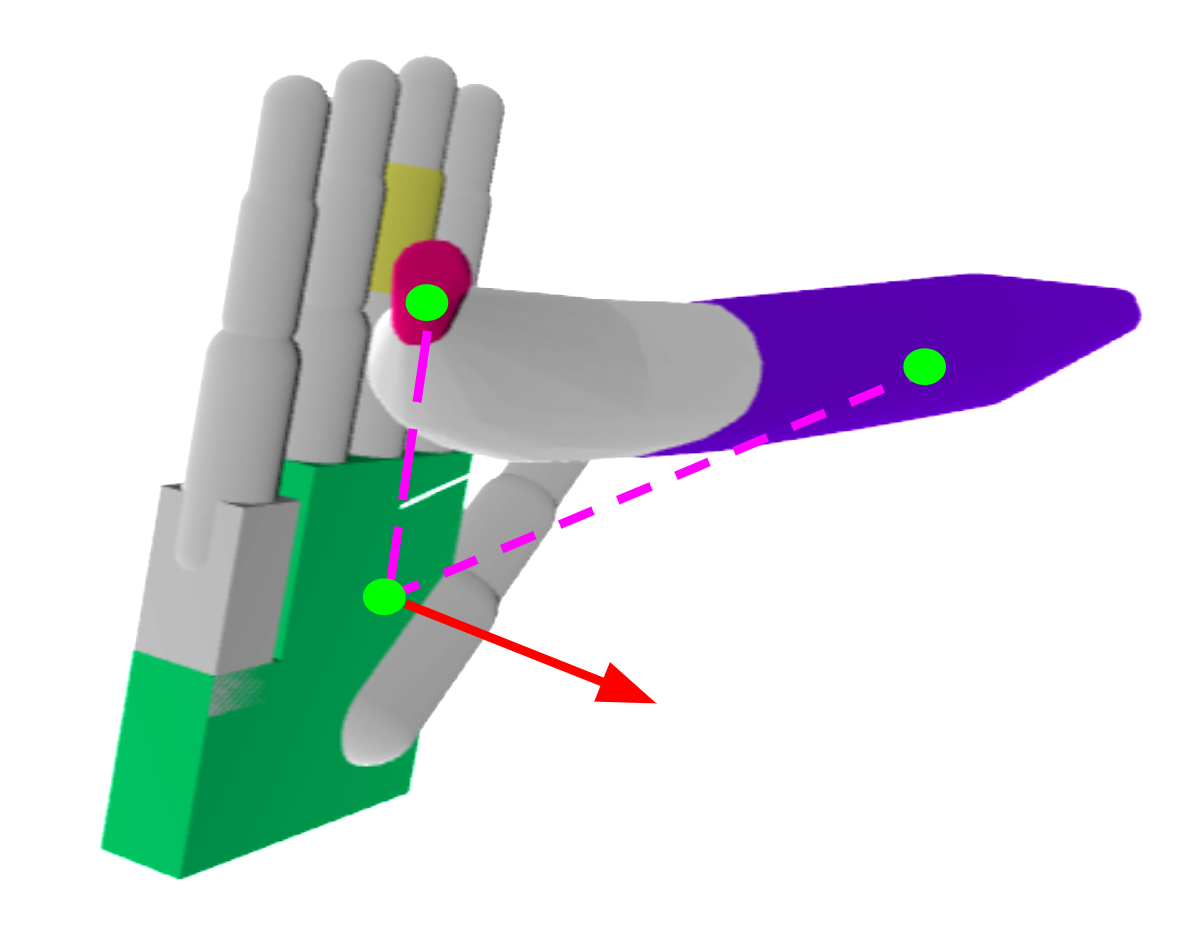}} \subfigure[]{\includegraphics[width=0.477\linewidth]{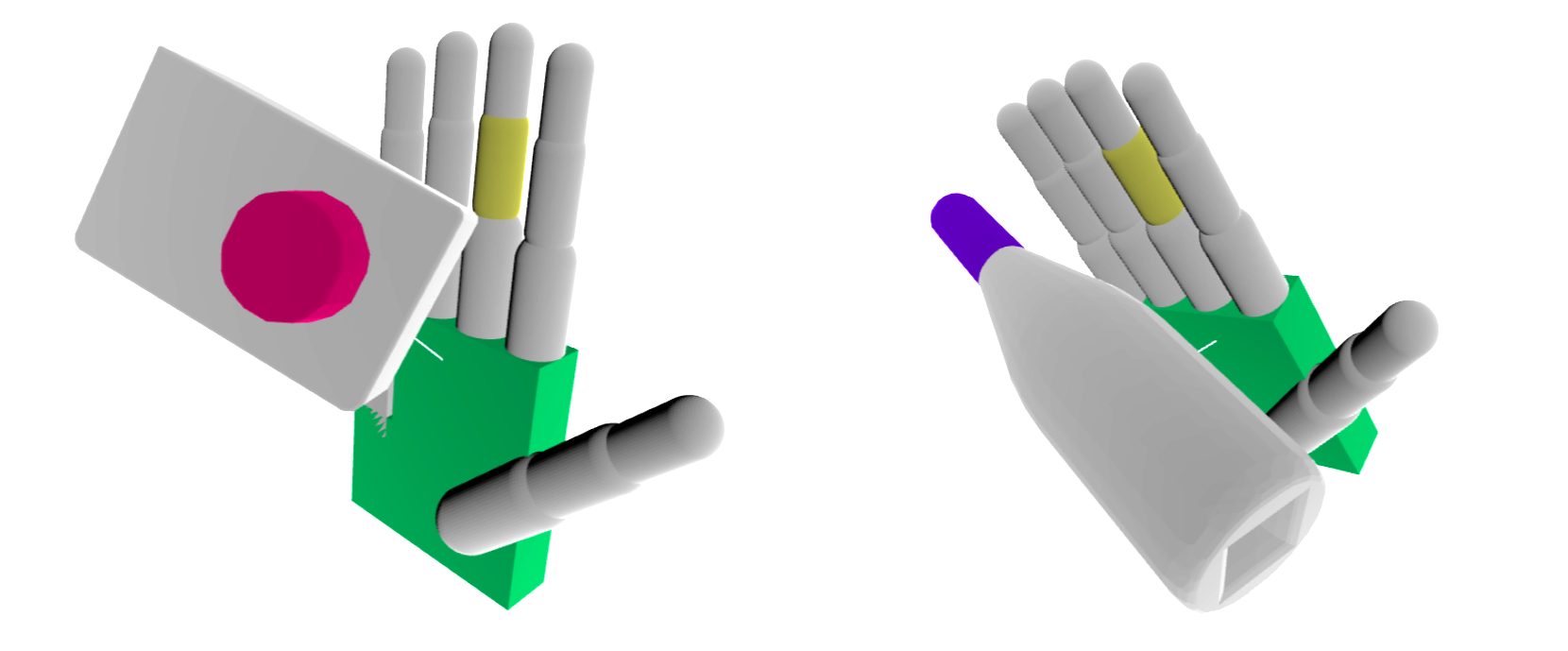}}
\caption{(a) Metrics of distances from the middle phalanx of the middle finger and the palm of the gripper's hand to the center of the object; (b) Metric of the angle between the directions to the edges of an oblong object and the vector of normal to the palm of the hand; (c) Examples of pose sampling for Bottle and Camera objects.}\label{ruki}
   %\vspace{-0.3cm}
\end{figure}

\begin{figure}[h!]
  \centering

  \subfigure[]{ \includegraphics[width=0.6\linewidth]{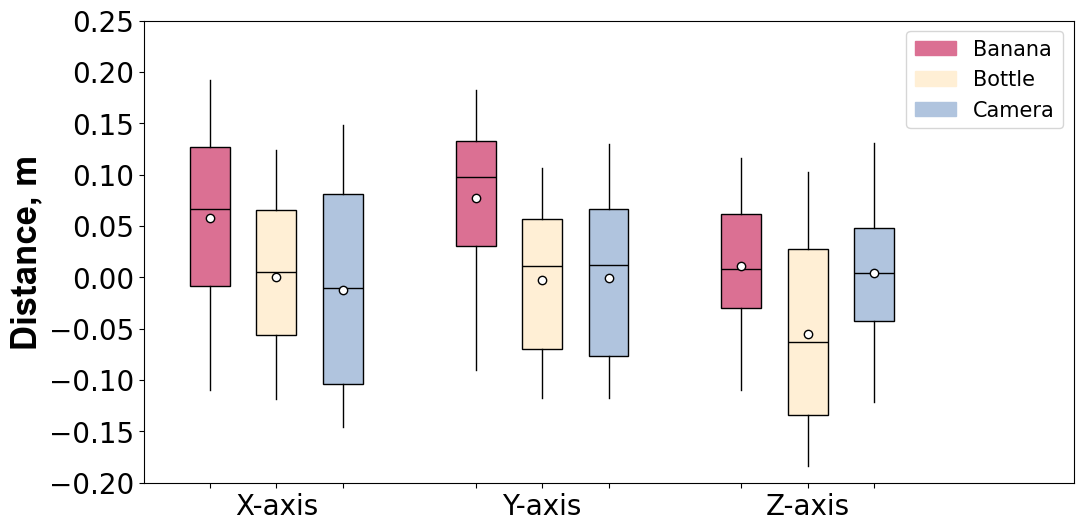}}
    \subfigure[]{\includegraphics[width=0.6\linewidth]{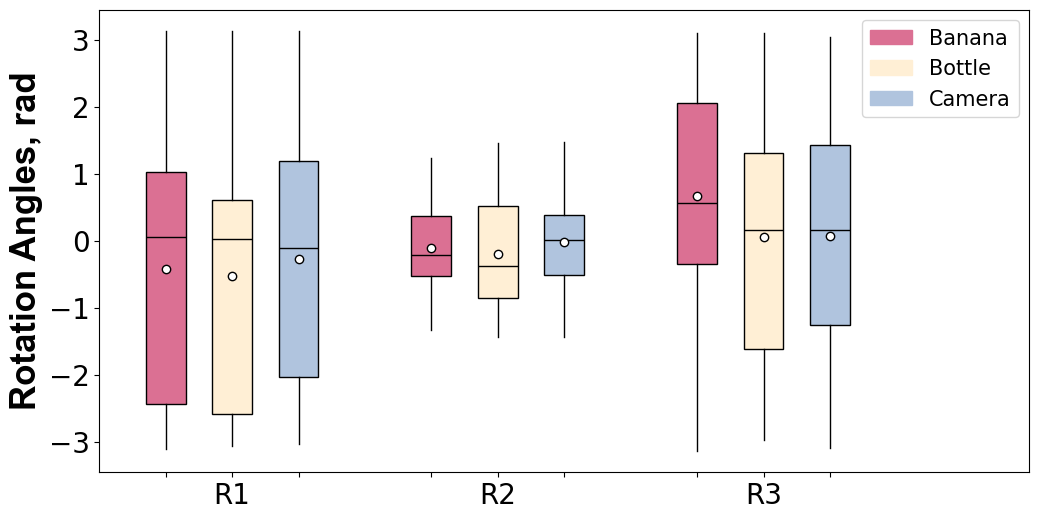}}
    \subfigure[]{\includegraphics[width=0.6\linewidth]{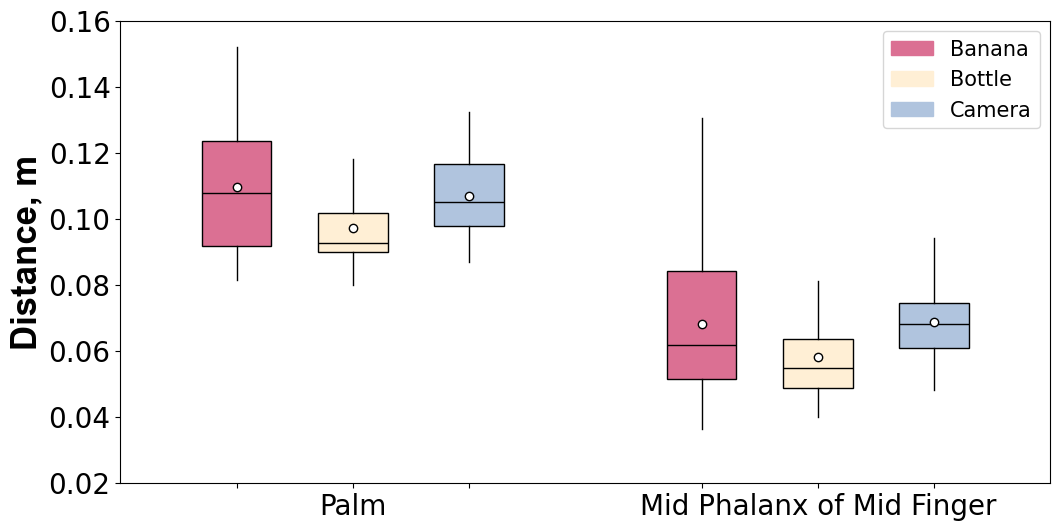}}
   \subfigure[]{\includegraphics[width=0.6\linewidth]{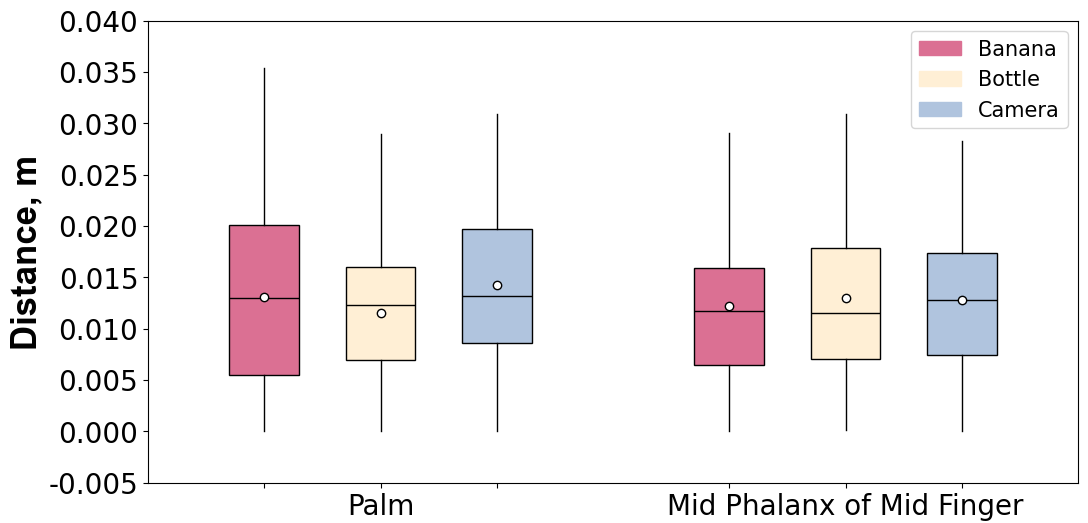}}
   \hfill
    \subfigure[]{\includegraphics[width=0.33\linewidth]{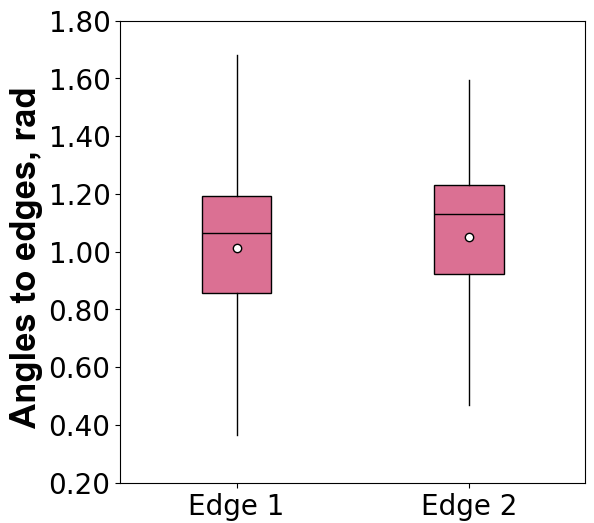}}
   \caption{ (a) Distribution of axial distances between the gripper and the object; (b) Distribution of angles of rotation of the gripper relative to the object; (c) Distances between the centers of the considered parts of the gripper and the object; (d) Minimum distances between any point of the considered parts of the gripper and the object; (e) Angles between the normal vector of the palm and the directions to the edges of the Banana object.}\label{fig:statistics_sampled}
   \vspace{-0.3cm}
\end{figure}

In order to evaluate the efficiency of training the diffusion model, an algorithm had to be implemented to generate poses of objects not included in the training dataset. These poses should be realistic and diverse. Since the DexGraspNet dataset contains such examples, an algorithm was built for sampling poses based on the statistics of the available data from it.
First, statistics were collected on the values of three coordinates and three Euler angles of rotation relative to the gripper for each of the three objects. Only samples from the dataset used to train the RL agent with a dynamically changing initial robot pose were considered. These statistics are shown in Fig. \ref{fig:statistics_sampled}(a, b).

Since random selection of coordinate and rotation values from the calculated distribution was unsuitable for pose generation, we introduced metrics for pose validation. To estimate the distance to a realistic grip area, we measured two values: the distance from the middle phalanx of the middle finger and from the center of the palm to the object (Fig.~\ref{ruki}(a), Fig.~\ref{fig:statistics_sampled}(c)). We also computed the minimum distances between all points of these considered ShadowHand parts and the object (Fig.~\ref{fig:statistics_sampled}(d)). For oblong Banana object, we added angular boundary metrics between the vectors pointing to its edges and the palm’s normal vector (Fig.~\ref{ruki}(b), Fig.~\ref{fig:statistics_sampled}(e)).

The values of coordinates and rotation angles for each object were sampled from a uniform distribution \(\mathcal{U}(a, b)\), where $a$ and $b$ are defined as follows:  
\[
a = \frac{Q_1 + L}{2}, \quad b = \frac{Q_3 + U}{2},
\]  
with \( L \) and \( U \) being the boxplot whiskers:  
\( L = Q_1 - 1.5 \times \text{IQR} \), \( U = Q_3 + 1.5 \times \text{IQR} \), \( Q_1 \) and \( Q_3 \) are the first and third quartiles (boxplot bounds), \( \text{IQR} = Q_3 - Q_1 \) is the interquartile range.

\begin{figure}[h!]
  \centering
     %\subfigure[Diffusion Model MSE Loss.]{\includegraphics[width=0.48\linewidth]{pictures_grasp/loss_diffusion.png}} 
    \subfigure[]{
    \includegraphics[width=0.55\linewidth]{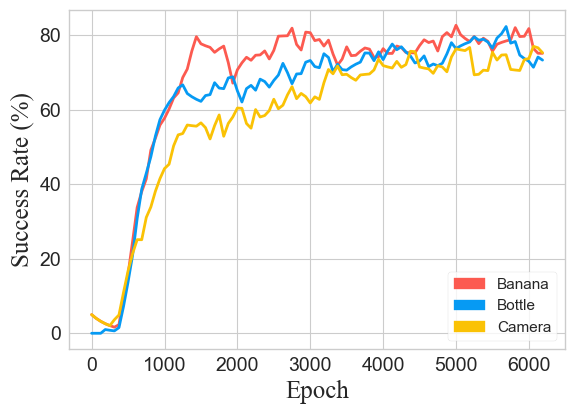}}
    \hfill
    \subfigure[]
    {\includegraphics[width=0.395\linewidth]{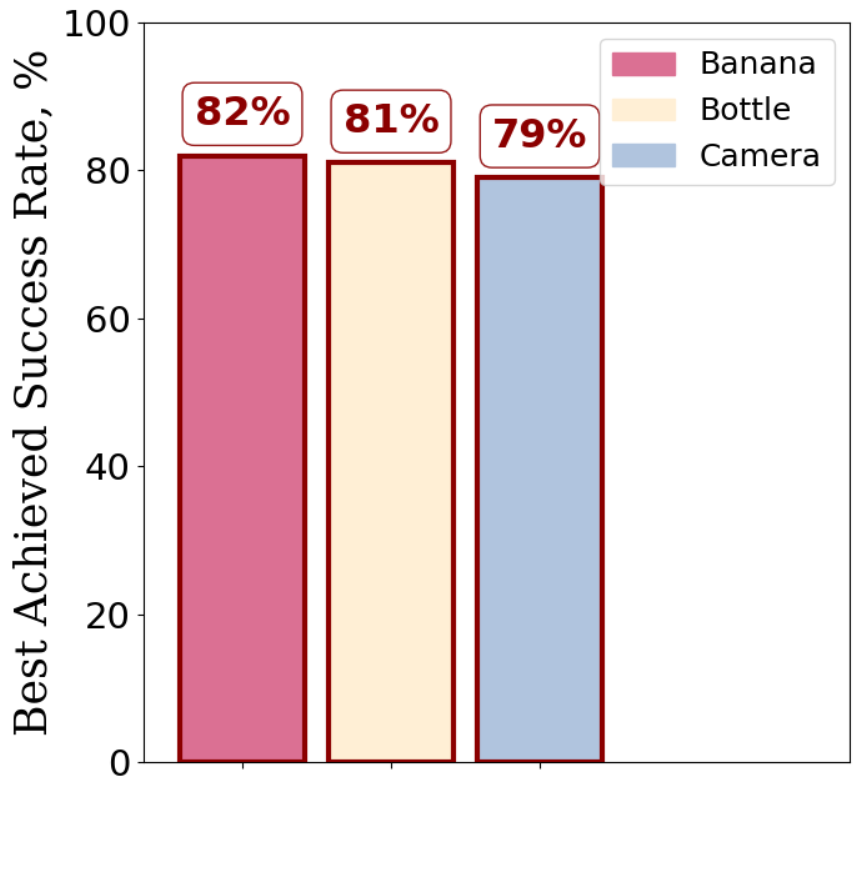} 
    }
   \caption{(a) Mean Success Rate of the Diffusion Model for random object poses; (b) Best Achieved Mean Success Rate on Random Poses. Mean Success Rate is a mean percentage of transitions that have a success reward during validation.}\label{fig:diffusion_metrics}
   \vspace{-0.15cm}
    
\end{figure}

The boundary values for the distances between the centers of the palm and middle phalanx of the middle finger to the object were calculated as \(D_{max} = Q_3 + 1.5 \times \text{IQR}\). The minimum allowable distances between any points of these parts of the ShadowHand and the object were calculated as \(D_{min}= Q_1 - 1.5 \times \text{IQR}\). For the Banana object, it was checked that the angles of directions to its edge parts lie in the range from \( \theta_{min}= Q_1 - 1.5 \times \text{IQR} \) to \( \theta_{max} = Q_3 + 1.5 \times \text{IQR} \).

 Finally, the sampled object poses were additionally checked for collisions with the robotic hand.
}

\subsection{Diffusion Model Training.}

After collecting the dataset and processing it as described in Section \ref{diffusion_theory}, the diffusion model was trained on three objects: Banana, Bottle and Camera. The number of training iterations for each of the objects was 100 000 with a batch size of 16. The mean square error (MSE) was chosen as the loss function, and the same reward function from Section \ref{rl_grasp_theory} was chosen for success validation. As a noise scheduler, we used the DDPM scheduler with a glide cosine scheduler (i.e., \texttt{squaredcos\_cap\_v2}) and a step number of 50. After each subsequent 1000 iterations, the model was validated on 50 episodes with a positioned object according to the pose sampling algorithm from Section \ref{sampling_grasp}. 

\begin{figure}[h!]
  \centering
\includegraphics[width=0.9\linewidth]{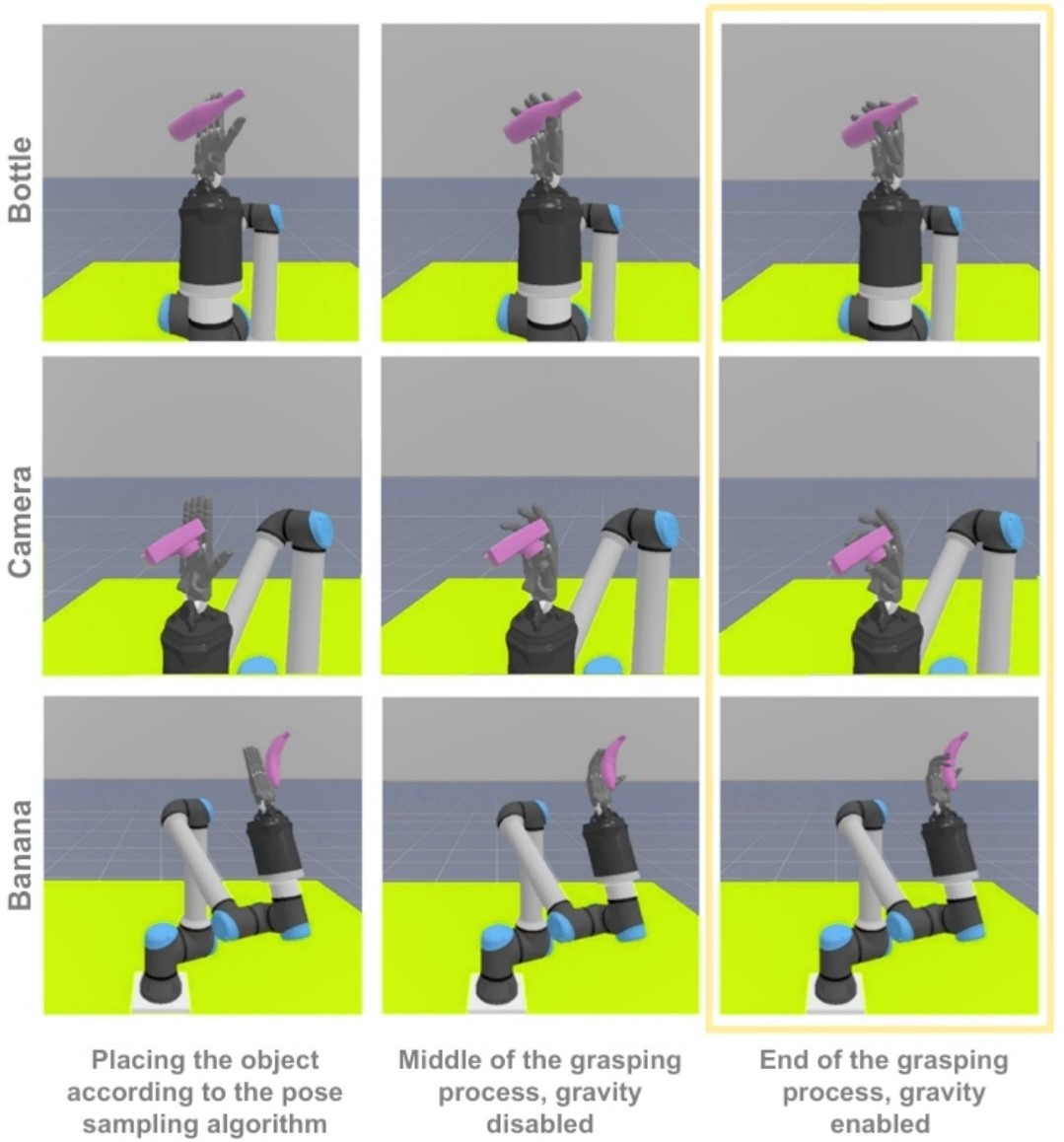}
   \caption{Example of grasping process with trained Diffusion Model.}\label{fig:diffusion_example}
\vspace{-0.35cm}
\end{figure}

The results of training the models can be seen in Fig. \ref{fig:diffusion_metrics}, which illustrates their convergence, as well as their high efficiency in capturing randomly located objects during validation. The Mean Success Rate is close to 80\%, which proves the applicability of the constructed pipeline for training a diffusion model to capture objects in complex unfamiliar scenarios.
An illustration of examples of actions predicted by the diffusion model for each of three objects can be seen in Fig. \ref{fig:diffusion_example}.

\section{Conclusion} 
In this paper, we demonstrate the efficacy of employing a lightweight RL agent to refine samples from a large pre-built dataset, thereby enhancing the quality, stability, and variability of grasp examples for subsequent diffusion policy training. The trained model has been shown to achieve a high success rate close to 80\% in grasping randomly placed objects, thereby demonstrating its robustness and generalization ability.

The proposed methodology has the potential to significantly reduce the necessity of manual data collection for the effective training of diffusion models for dexterous multi-finger grasp manipulation. It can also be useful for training auxiliary VLA diffusion models, such as the one described in the aforementioned work \cite{grasp_118_main}. With our pipeline, it is now possible to train such diffusion models without collecting datasets for each specific scenario. Instead, existing datasets can be adapted in a simulation environment using an RL agent. The pose sampling algorithm can also be used to increase the number of samples in a dataset and improve their diversity.

Future plans include preparing a large dataset, adding more multimodal data for training RL agents, integrating the approach with Vision-Language-Action (VLA) models, and validation in real-world conditions.
Due to its flexibility, the pipeline can be easily integrated to train various robotic tasks in different environments based on pre-built datasets, simplifying the task of collecting data for training diffusion policies for practical applications.

\bibliographystyle{IEEEtran}
\bibliography{exampl}  % .bib

\end{document}